# An Effective Networks Intrusion Detection Approach Based on Hybrid Harris Hawks and Multi-Layer Perceptron

Moutaz Alazab[a], Ruba Abu Khurma[b], Pedro A. Castillo[c], Bilal Abu-Salih[d], Alejandro Martín[f], David Camacho[f]

[a]Faculty of Artificial Intelligence, Al-Balqa Applied University, Amman, Jordan m.alazab@bau.edu.jo
[b] Computer Science Department, Faculty of MEU Research Unit, Middle East University, Amman, Jordan. rubaabukhurma82@gmail.com,
[c] Department of Computer Engineering, Automation and Robotics, University of Granada, Granada, Spain pacv@ugr.es,
[d] King Abdullah II School of Information Technology, The University of Jordan, Amman, Jordan b.AbuSalih@ju.edu.jo,
[e] Department of Computer Systems Engineering Universidad Politécnica de Madrid, Madrid, Spain, alejandro.martin@upm.es, david.camacho@upm.es,
[f] Computer Science Department, Faculty of MEU Research Unit, Middle East University, Amman, Jordan. rubaabukhurma82@gmail.com,

**Abstract**

This paper proposes an Intrusion Detection System (IDS) employing the Harris Hawks Optimization algorithm (HHO) to optimize Multilayer Perceptron learning by optimizing bias and weight parameters. HHO-MLP aims to select optimal parameters in its learning process to minimize intrusion detection errors in networks. HHO-MLP has been implemented using EvoloPy NN framework, an open-source Python tool specialized for training MLPs using evolutionary algorithms. For purposes of comparing the HHO model against other evolutionary methodologies currently available, specificity and sensitivity measures, accuracy measures, and mse and rmse measures have been calculated using KDD datasets. Experiments have demonstrated the HHO MLP method is effective at identifying malicious patterns. HHO-MLP has been tested against evolutionary algorithms like Butterfly Optimization Algorithm (BOA), Grasshopper Optimization Algorithms (GOA), and Black Widow Optimizations (BOW), with validation by Random Forest (RF), XG-Boost. HHO-MLP showed superior performance by attaining top scores with abstractabstractaccuracy rate of 93.17%, sensitivity level of 89.25%, and specificity percentage of 95.41%.

*Keywords:* Multi-layer Perceptron (MLP), Harris Hawks Optimization (HHO), Intrusion Detection System (IDS)

## 1. Introduction

Computer networks are a group of interconnected nodes that are distributed within a local or wide geographic area to allow end-users to transmit and receive data over a communication medium (wired or wireless). The primary goals of building a computer network are to share resources (hardware, software, or data), to communicate between remote users using (digital audio, digital video, or text), and to provide various types of services such as web services (the World Wide Web) and application services (databases), and communication services (social networks). Recently, there has been remarkable progress in the networking field by designing different types of networks that differ based on several criteria such as typologies, protocols, architectures, and size [1].

One of the serious problems that may arise in computer networks is security and privacy breaches [2]. Digitization facilitates the work of hackers to carry out their criminal missions and cause security disasters. Cyber attackers take advantage of weaknesses within a network to infiltrate and cause disruptions or even bring it down altogether. A distributed denial-of-service (DDoS) attacks are a common security threat. They involve flooding a server with fake requests to clog up network channels and block legitimate requests [3]. Other security problems may occur by running malicious code on a server that changes or disrupts the functionality of the network [4].

Intrusion or unauthorized access takes place when network users exceed the privileges assigned to them [5]. Therefore there are a set of rules and practices that prevent any illegal access to the network. Intrusion detection systems (IDS) are typically implemented using a software-driven method. They identify abnormal behavior in a network and pinpoint evidence of security breaches. This system is crucial in protecting digital environments against potential threats and unauthorized entry. Signature and anomaly detection methods are the two main categories of IDS [6]. Both methods rely on analyzing network traffic to detect malicious patterns. However, the main difference is in the detection process. Signature-based methods detect mali-cious patterns like malware, while anomaly-based detection methods monitor any deviation from normal activity [7].

Anomaly detection methods typically use machine learning algorithms to increase network security [8, 9]. This depends on some features that help the algorithm distinguish regular traffic from malicious ones. Recent investigations have utilized various machine learning techniques to create efficient network intrusion detection systems [10]. Meta-heuristic techniques are often used in network IDS design to reduce the discrepancy between malicious and legitimate traffic detection. [11].

Swarm-based algorithms are meta-heuristic algorithms that simulate the natural survival of animals in nature [1, 12]. They are based on solid mathematical methodologies that reflect the social relations of animals that live in groups such as the colonies of bees, flocks of birds, and swarms of wolves. Swarm-based algorithms have proved their efficiency in solving various optimization problems. HHO is a swarming algorithm that inspires the hunting mechanism of Harris's Hawks when they find and pounce on their prey.

HHO has been used in different applications for solving several optimization problems [13, 14, 15]. The main reason behind selecting HHO in this work is that it has a suitable opportunity to utilize its features to build a reliable secure network IDS. The main features of HHO that encouraged using it for optimizing network security are: that it can effectively balance between exploring the search regions and exploiting them using a single parameter that controls the energy of the Hawks. It also uses an adaptive update strategy, which changes the position of the solutions in the search area in a time-varying fashion [16]. This allows the number of best solutions to decrease relative to increasing the number of iterations. HHO also has four stages of exploitation which enhance the local search and help the optimizer overcome some search problems such as premature convergence.

The suggested solution makes use of the HHO to enhance network-based IDS's ability to detect malicious traffic by choosing the best multi-layer perceptron parameters [17]. The selection of MLP over other classification methods is because MLP is a simple structure neural network. This means that it can produce accurate detection results within a promising running time compared with other complex deep-learning methods. Detection time is an important factor that a researcher must consider when designing a network IDS.

The following is how the paper is set up: Some of the earlier relevant studies are described in section 2 The specifics of the HHO and MLP algorithms are presented in section 3. The new evolutionary IDS is described in section 4. Section 5 displays the experimental methodology and the outcomes. The key conclusions and the future directions of the work are summarized in section 6.

## 2. Literature review

This section presents some of the latest research that proposed evolutionary-based methodologies for building promising and reliable security-aware network IDS. For more information about evolutionary-based IDS systems, a reader can refer to the survey [18].

Pozi et al. (2016) [19] introduced an innovative technique for detecting rare attacks by combining Support Vector Machines (SVMs) and Genetic Programming (GP). Their proposed GPSVM method demonstrated its remarkable capability of accurately detecting rare and anomalous attacks with increased precision. The study's findings showed that GPSVM maintained comparable levels of accuracy while achieving a greater detection rate for uncommon attacks. More specifically, GPSVM detected DoS attacks with an outstanding 94.56% geometric mean detection rate. Notably, this study did not employ feature selection or resampling techniques; its findings demonstrate the efficiency of the GPSVM approach in rare attack detection. Furthermore, this method could potentially enhance the accuracy and reliability of intrusion detection systems while strengthening overall network security.

Ali [20] developed a PSO FLN IDS model, which is based on the principles and techniques of Particle Swarm Optimization (PSOs) and Fast Learning Networks. This model's primary objective is to optimize values for neurons in hidden layers. The PSO FLN model, which achieved an impressive accuracy of 99.68% during testing was developed to combat the problem of decreased precision for certain categories due to a lack of training data.

The authors in [21] conducted an innovative network intrusion detection system by employing the Firefly algorithm (FA) to improve K-means clustering efficiency. Their proposed approach was tested against six other clustering methodologies such as K-Means enhanced by Cuckoo, K-Means with Bat, K-Means++, Canopy, and Farthest First; with its results showing it outperforming traditional classification techniques by attaining an outstanding recall rate of 72.6%.

Chen et al. [22] described how an IDS was developed using fuzzy clustering techniques implemented within a cloud computing infrastructure, using fuzzy clustering as its foundation. Experimental results demonstrated this approach was capable of considering 10 attack types simultaneously while outperforming alternative models with an accuracy rate of 80

Mohammadi (2019) [23] developed an IDS by employing clustering techniques. His method involved combining cuttlefish algorithms and Decision Trees (DT) for optimal system performance; his results achieved remarkable accuracy (95.03%) and detection rates (95.23%), with minimal false-positive rates (1.65%). These performances outshone any established methods found in scholarly works.

Kalaivani [24] developed an effective IDS utilizing the Artificial Bee Colony (ABC) algorithm for cloud environments. He successfully employed his proposed classification model with 96% accuracy - far surpassing other methods. Further work will include improving existing classifiers as well as creating hybridized classification systems with even greater performance potential.

Ren et al. [25] conducted an innovative IDS by combining SVM, DT, and Genetic Algorithm (GA). SVM served as the learning mechanism, while DT provided feature selection functionality, and GA provided optimization techniques to improve FS processes. Comparable with existing algorithms, this integrated model showed remarkable proficiency at identifying infrequent anomalous behaviors with an impressive accuracy rate of 93.55%. This advanced system could potentially have applications in other areas of anomaly detection such as fraudulent activities; however, the training process for classifiers may be time-consuming, suggesting there could be additional optimization of search strategies.

Elhag [26], designed a fuzzy network IDS approach that was based on using a multi-objective evolutionary algorithm which allowed the users to select the solutions that are best suited for the network features. It achieved (accuracy=98.10%). Comparing the proposed method with FARC-HDclassifiers, FARCHD with OVO, and C4.5 decision trees, the high quality of this methodology was demonstrated. The proposed method achieved a good balance between Precision and interpretability in all cases.

Benmessahe [27], developed a reliable network IDS called (FNN-LSO-IDS) based on Locust Swarm Optimization (LSO) and Feed-forward Neural Network (FNN). This method improves the detection rate and convergence speed, as well as reliability, due to a reduced chance of being caught in local minima.

The authors in [28] presented an innovative method for detecting abnormal traffic using an RNN implemented within the Apache Spark Framework.

The authors reported outstanding IDS evaluation metrics, such as detection accuracy, detection rate, and false alarm rate, for their UNSW-NB15 dataset, with scores of 95.42%, 99.33%, and 9.40% achieving detection accuracy; 94.02% 89.83% and 2.21% were achieved with NSL-KDD dataset respectively. These scores outshone those obtained through other training techniques, demonstrating the promise of their ENN (FNN-LSO) approach for creating practical IDSs. However, experiments were performed only on subsets of datasets; therefore the authors acknowledged that additional research with larger datasets and more powerful hardware infrastructure is required to confirm their approach's efficacy. Still, their results provided promising directions for future IDS research.

Naik [29], proposed a technique based on Teaching-Learning Based Optimization (TLBO), the Functional Link Neural Nets (FLANN), mutation operation, and elitism to build a reliable IDS that can deliver accurate security results. Mutation operations provide an efficient solution for handling redundant parameters, avoiding palindrome occurrences while also dramatically increasing the efficiency of the method. By decreasing computational load, the proposed technique shows an impressive 96.30% detection rate.

Almomani et al. (2020) [30], presented a feature-based technique for network IDS. This study incorporated four meta-heuristic advanced techniques: Grey Wolf Optimization (GWO), Particle Swarm Optimization (PSO), Firefly Algorithm (FA), and Genetic Algorithm (GA). These strategies were used to improve the performance of IDS. This multi-faceted approach, which harnessed the strengths of every algorithm, aimed to enhance the IDS's capabilities in identifying potential security threats and mitigating them. By employing wrapper methods in combination with MI filter methods, this system was able to select features effectively for intrusion detection. With using the J48 classifier as part of its proposed methodology, classification accuracy rates reached between 79.175% and 90.484% when employing classification accuracy rates between 80.1750%-90.5484%. Applying SVM classifiers yielded classification accuracy rates ranging from 79.077% to 90.119%. These results demonstrate the success of feature selection processes by showing that most feature reduction rules in proposed models outperformed those that utilized all available features, demonstrating their efficacy as part of a selection process. They advise that, despite positive results obtained, further advanced learning structures like Recurrent Neural Nets (RNNs) and Convolutional Neuronal Networks be explored for comprehensive evaluation of model performance and efficiency.

In [9] a new approach was proposed based on enhancing the MFO by adopting new operators besides the embedded spiral operator to balance the exploration and exploitation alleviating the local minima problem. The main contribution of this work is the adoption of the cosine similarity measure to binarize the continuous MFO into a binary problem. Cosine similarity overcomes the limitations of the commonly used sigmoid function that depends on using a threshold value for conversion. However, cosine similarity computes the similarity ratio between the current solution and the optimal solution. The augmented MFO wrapper framework was applied as an IDS to detect anomalous traffic in the network. The proposed method was compared against several well-known state-of-the-art algorithms on three network datasets (KDDCUPP9, NSL-KDD, and UNSW-NB15), using IDSACC, IDSTPR, IDSFPR, IDSF-score, and convergence evaluation measures to assess the performance of the proposed method. The experimental results demonstrated the superiority of the proposed cosine similarity method compared to other algorithms with an accuracy of 97.8%, F-score of 99%, TPR of 99.6%, and FPR of 8.1% using only five selected features from the KDDCUPP99 dataset. It achieved the accuracy of 89.7%, TPR of 89.1%, FPR of 2.9%, when four selected features from the NSL-KDD dataset are used. And finally, it achieved an accuracy of 92.4%, TPR of 92.3%, FPR of 3%, and F-score 94.2% when the UNSW-NB15 dataset is used.

In [31], the authors proposed a hybridization of modified binary GWO and PSO. The proposed solution used two benchmarking datasets, NSL KDD'99 and UNSW-NB15, and the results revealed that the proposed solution outperformed the existing solutions, as the proposed approach improved the detection accuracy by approximately 0.3% to 12%, and the detection rate by 2% to 12%. In addition, it reduces false alarm rates by 4% to 43%, and reduced the number of features by approximately 31% to 75%. Last, the proposed approach reduced processing time by approximately 14% to 22% compared to state-of-that-art approaches.

Overall, machine learning techniques have a major impact on designing IDS that are capable of improving the security of different types of networks. This idea can be used by researchers to continue this line of research by proposing different optimization algorithms, new enhancement operators, and novel updates to be evaluated and tested on different network datasets and in different processing platforms. This opens the opportunity for cybersecurity researchers to be up to date and provide state-of-the-art solutions capable of addressing new challenging security breaches emerging in networks. The main positives of the proposed methodology, is to simultaneously optimize the number of selected features in addition to optimizing the parameters of the MLP neural network. Optimizing the feature set and the parameters of the MLP can generate a better model for classifying the malicious and normal traffic. So that the classification results produce fewer errors and better performance accuracy. The most recent HHO algorithm in particular was chosen because of its outstanding attributes for adaptive convergence and balancing the exploration and exploitation stages of search results. This significantly improves the classification performance of the optimizer and helps alleviate local minima. This research's main contributions can be summed up as follows:

- Introduction and implementation of the novel HHO-MLP approach. In this approach, HHO serves as an initial preprocessing step which helps accelerate the FS process and ensures MLP learns about relevant and informative network traffic features. HHO can also optimally assign weight and bias values for improved model performance. MLP uses multiple layers of hidden nodes instead of relying on just one or two to obtain more accurate results compared to traditional methods. This innovative architecture is intended to produce more precise outcomes compared to their counterparts.

- Utilization of an extensive set of features and instances within network intrusion detection datasets enables models to develop a deeper understanding of their data, leading to improved predictions.

## 3. The proposed detection methodology

We examine the research methods used in our investigation in this section of the paper. The HHO algorithm and MLP are the two main parts of this strategy. We'll go over each of these separately.

*3.1. HHO Algorithm*

HHO inspires Harris's Hawks strategy to hunt and capture prey in the environment. Heidari [32] developed the HHO methodology in such a way there are two exploration strategies and four exploitation strategies.

Fig. 1 depicts the two approaches employed by HHO.

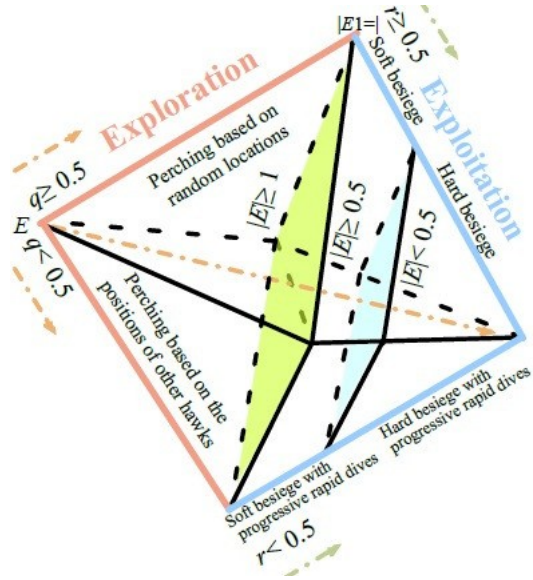

Figure 1: HHO exploration and exploitation strategies

### 3.1.1. HHO exploration strategies

HHO applies two strategies for perching based on the value of a random variable q. When q ≥ 0.5 in, the first technique is implemented. Eq. (1) and the second technique is used when q ≥ 0.5 in Eq. (1).

$$H(iter+1) = \begin{cases} H_{random}(iter) - r_1 H_{random}(iter) - 2r_2 H(iter), & q \geq 0.5 \\ (H_{prey}(iter) - H_m(iter)) - r_3(LoBo + r_4(UpBo - LoBo)), & q < 0.5 \end{cases} \quad (1)$$

where the location of the Hawks in the following cycle is H(iter + 1) Prey's location is iter, $H_{prey}(iter)$, and the solutions' current location is H(iter). In the range of (0,1), $r_1$, $r_2$, $r_3$, $r_4$, and q are all random values. The variables' highest and lowest values are shown by LoBo and UpBo, respectively. The random solution is called $H_{random}(iter)$, and the average position of the current swarm of solutions is called $H_m$. The average hawk location is calculated using Eq. (2):

$$H_m(iter) = \frac{1}{N} \sum_{i=1}^{N} H_i(iter) \quad (2)$$

where N is the total number of solutions, and $H_i(iter)$ is the position of each solution in a given iter.

HHO alternates between the global and local search phases before moving between the various exploitation phases. Prey energy is calculated as follows:

$$E = 2E_0(1 - \frac{iter}{all\text{-}iter}) \tag{3}$$

where E is the power of the prey, all-iter is the total number of cycles, and $E_0$ is the starting energy that randomly changes in (-1, 1) at each cycle.

*3.1.2. HHO exploitation strategies*

In the HHO, there are four potential exploitation methods. Assume that the chance for a prey to successfully escape is (r <0.5) and that the chance of a prey being unsuccessful in escaping is (r ≥0.5).

- SB: r ≥ 0.5 and E ≥ 0.5.

$$H(iter + 1) = \Delta(H(iter) - EJH_{prey}(iter) - H(iter)) \tag{4}$$

$$\Delta H(iter) = H_{prey}(iter) - H(iter) \tag{5}$$

where $\Delta H(iter)$ is the difference between the prey and the current location in cycle iter, $r_5$ is a random value in (0,1), and $J = 2(1 - r_5)$ is the random jump strength of the prey. The J value changes randomly in each cycle.

- HB: r ≥0.5 and E <0.5. Eq. (6):

$$H(iter + 1) = H_{prey}(iter) - E\Delta H(iter) \tag{6}$$

- SB-PRD: E ≥0.5 but r <0.5. The levy flight (LeFl) is applied in Eq. (7):

$$Y = H_{prey}(iter) - EJH_{prey}(iter) - H(iter) \tag{7}$$

The LeFl is applied as in Eq. (8).

$$Z = Y + Size \times LeFl(Dim) \tag{8}$$

where Dim is the dimensionality of the problem and Size is a random vector by size 1× Dim and LeFl is the levy flight function, which is shown in Eq. (9):

$$LeFl(x) = 0.01 \times \frac{u \times \sigma}{v^{\frac{1}{\beta}}}, \sigma = (\frac{\Gamma(1+\beta) \times \sin(\frac{\pi\beta}{2})}{\Gamma(\frac{1+\beta}{2}) \times \beta \times 2^{(\frac{\beta-1}{2})}})^{\frac{1}{\beta}} \tag{9}$$

where u and v are two arbitrary quantities that lie within an open interval (0,1); these random variables may take any value between 0 and 1, including endpoints. Assume $\beta$ remains constant by assigning it a numerical value of 1.5. SB strategy can be applied as in Eq. (10):

$$H(\text{iter} + 1) = \begin{cases} Y & \text{if } F(Y) < F(H(\text{iter})) \\ Z & \text{if } F(Z) < F(H(\text{iter})) \end{cases} \quad (10)$$

where Y and Z are obtained using Eq. (7) and Eq. (8).

- HB-PRD

When $|E| < 0.5$ and $r < 0.5$, the following equation is applied:

$$H(\text{iter} + 1) = \begin{cases} Y & \text{if } F(Y) < F(H(\text{iter})) \\ Z & \text{if } F(Z) < F(H(\text{iter})) \end{cases} \quad (11)$$

where Y and Z are computed in Eq. (12) and Eq. (13).

$$Y = H_{\text{prey}}(\text{iter}) - EJH_{\text{prey}}(\text{iter}) - H_m(\text{iter}) \quad (12)$$

$$Z = Y + \text{Size} \times \text{LeFl}(\text{Dim}) \quad (13)$$

where $H_m(\text{iter})$ is computed in Eq. (2).

### 3.1.3. Pseudo-code of HHO

HHO algorithm can be represented using Algorithm B as shown in Algorithm 1.

### 3.2. MLP architecture

The Artificial neural network (ANN) has been commonly used as a learning algorithm to perform a training process on a given data instance and generate a pattern (data model) that is used then to predict the output of another hidden part of the dataset in the testing process. The MLP connects the neurons in the hidden layer with n weights and one bias [33, 34]. Each hidden neuron performs two primary functions: the summation as illustrated in Eq. (14) and the activation as illustrated in Eq. (15). Neuron j that performs a summation operation then uses its activation function to transform its output, producing results which are further used for computation within its neural network. Each neuron's functionality in a neural network depends on two key processes - summation function and activation function - both combining input signals from various neurons into aggregate sums that are then transformed by activation into specific output values that can then be further processed through subsequent computations within its network.

**Algorithm 1** HHO pseudo-code
---
**Inputs**: The total number of cycles (T) and the swarm size (N)
**Outputs**: The prey's location and fitness worth
random initialization of the swarm $H_i(i = 1, 2, \ldots, N)$
**while** (there have been fewer iterations than T) **do**
    Calculate the solutions' fitness values.
    Modify the prey's location to $H_{prey}$
    **for** (every ($H_i$)) **do**
        alter $E_0$ and J         ▷ $E_0$=2random()-1, J=2(1-random())
        alter the E using Eq. (3)
        **if** ($|E| \geq 1$) **then**
            alter the hawk's position by Eq. (1)
        **if** ($|E| < 1$) **then**
            **if** ($r \geq 0.5$ and $|E| \geq 0.5$) **then**
                alter the hawk's position by Eq. (4)
            **else if** ($r \geq 0.5$ and $|Energy| < 0.5$) **then**
                alter the hawk's position by Eq. (6)
            **else if** ($r < 0.5$ and $|E| \geq 0.5$) **then**
                alter the hawk's position by Eq. (10)
            **else if** ($r < 0.5$ and $|E| < 0.5$) **then**
                alter the hawk's position by Eq. (11)
**Return** $H_{prey}$

$$\text{Sum}_j = \sum_{i=1}^{n} w_{ij} \times I_i + b_j \qquad (14)$$

Where $w_{ij}$ stands for the weight between the nodes of the input and hidden layers, respectively, and $b_j$ stands for the bias in favor of the hidden node, respectively.

$$y_j = \text{func}(\text{sum}_j) \qquad (15)$$

where $y_j$ represents the output neuron j, j = 1, 2, ..., m, and func, as stated in Eq.(16).

$$\text{func}(\text{sum}_j) = \frac{1}{1 + e^{-\text{sum}_j}} \qquad (16)$$

Using the summing and activation functions as stated in EEq. (17) and Eq. (18), the final outputs $Y_j$ are calculated based on the outputs of all hidden neurons.

$$\text{Sum}_j = \sum_{i=1}^{n} w_{ij} \times y_j + b_j \qquad (17)$$

where $b_j$ is the bias of the output neuron $j$, and $w_{ij}$ is the weight between the hidden neuron $i$ and the neuron $j$ in the output layer.

$$Y_j = \text{func}(\text{sum}_j) \qquad (18)$$

where the same sigmoid function as in Eq. (16) is utilized in func, $Y_j$, j = 1, 2, ..., k, and $Y_j$ is the final output j.

## 4. System Design and Implementation of the Proposed HHO-MLP

The proposed HHO-MLP relies on reducing the error in predicting network intrusions. To reduce the output error, the proposed HHO-MLP applies several steps as follows:

- One-dimensional arrays that represent the results of the HHO algorithm are used to encode the weights and biases of the MLP.

- The fitness function that calculates the intrusion detection error rate in each iteration of the optimization process is used to evaluate each member of the HHO swarm.

- An adaptive update strategy is implemented at each iteration to change the position of the swarm solutions and facilitate the switching between searching globally and locally.

- The HHO converges in the latest stages towards the best solution that represents the optimal weights and biases. These values of weights and biases are then used to build the architecture of the MLP.

Fig. 2 shows the steps of proposed HHO-MLP steps for performing intrusion detection in networks. Based on the flowchart, the first step is to encode the MLP in the HHO.

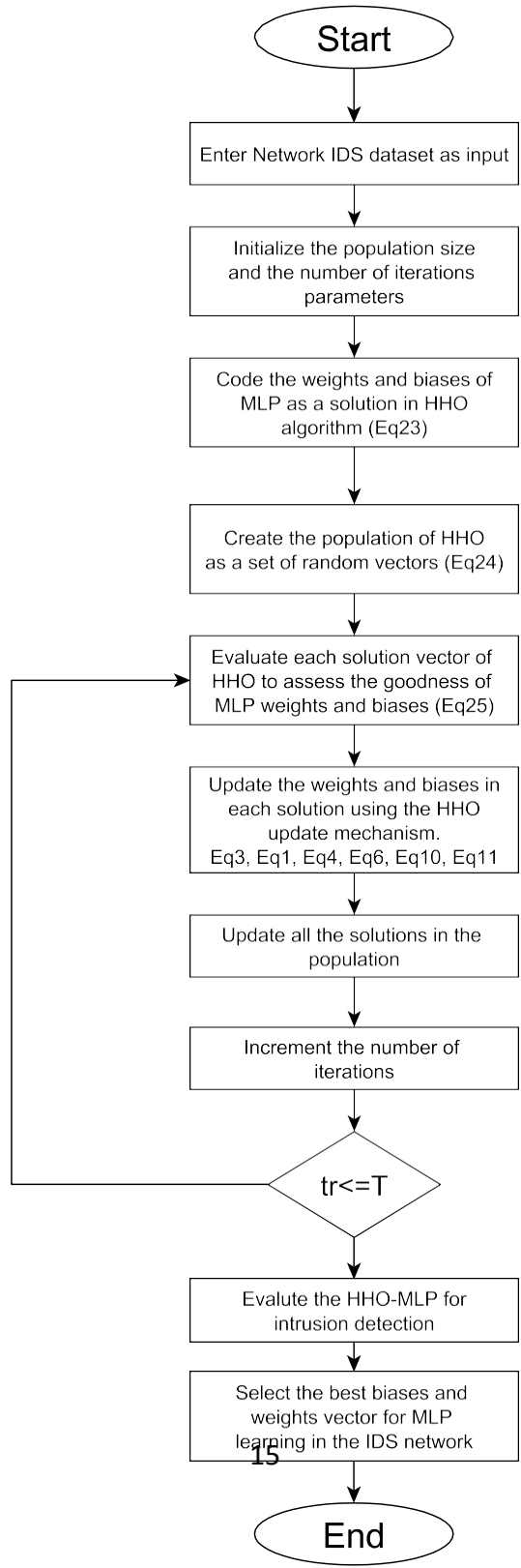

Figure 2: Flowchart of the HHO-MLP.

Two subprocesses make up the proposed HHO-MLP's preprocessing stage: data normalization and feature selection-based dimensionality reduction. The network data needs to be reduced and standardized before training. This refers to putting all network traffic feature values inside a certain range, as [a,b]. Eq. (19) shows the normalization equation.

$$\bar{f\,r} = \frac{fr - \text{min-traf}}{\text{max-traf} - \text{min-traf}}(Nb - Na) + Na \qquad (19)$$

where $\bar{f\,r}$ is the normalized value, $fr$ is the abnormalized value of the traffic features, min-traf and max-traf, are the lower and upper bound of the traffic, $Nb, Na$ are the limits of the normalization range. If the normalization range is in the [0,1], then the normalization of the equation is shown in Eq. (20)

$$\bar{f\,r} = \frac{fr - \text{min-traf}}{\text{max-traf} - \text{min-traf}} \qquad (20)$$

Normalizing the network traffic helps increase the classification accuracy. The next preprocessing step is dimensionality reduction which is based on implementing the feature selection process. The feature selection is carried out using the HHO algorithm. To represent the feature vector, a solution in the HHO algorithm's population is employed. The value of '0' for the feature vector means that the associated feature is not chosen whereas the value '1' means that the associated feature is chosen. Eq. (21) shows the feature vector of the network traffic that represents a Harris hawk or a solution in the population of the HHO algorithm.

$$HHO_i = [Fr_i^1, Fr_i^2, Fr_i^3, ..., Fr_i^D] \qquad (21)$$

$HHO_i$ is a Harris hawk or a solution in the swarm of the HHO algorithm. This represents a feature vector in the network traffic. $Fr_i^j$ is the jth element of the feature array $i$ and $D$ is the dimensionality of the problem. As mentioned previously, the values of the feature vector are either '1' or '0' means whether or not the feature is chosen. A set of feature vectors composes the population of candidate solutions which are initiated randomly as several harris hawks. The proposed methodology seeks to reduce both the number of attributes derived from network traffic as well as reduce error rates associated with intrusion detection. The basic goal is to use the fewest network traffic attributes possible to reduce intrusion detection. Therefore, Eq. (22)

shows the cost function used in the HHO algorithm to optimize the feature vector.

$$\text{Cost-IDS} = \alpha \times \text{Err} + \beta \times \frac{fr}{Fr} \qquad (22)$$

The Cost-IDS is the value associated with each Harris hawk during the optimization process. The aim is to find a solution with minimum Cost-IDS value. $\text{Err}$ is the error rate of intrusion detection. $fr$ is the chosen feature and $Fr$ is the set of all features. By applying the Hybrid Harmony Optimization and Multi-Layer Perceptron (HHO-MLP) approach to feature vectors, we can efficiently identify an optimal feature vector with minimum features while still maintaining an acceptable error rate for intrusion detection systems. These streamline feature selection while decreasing false alarm rates. This process further strengthens their performance and is an invaluable asset when applied correctly. The reduced features set is used then to train the MLP. Furthermore, the HHO-MLP can be used to enhance the weights and biases to minimize the intrusion error rate.

Eq. (23) demonstrates the set of biases and weights that a solution vector represents.

$$S(i) = \{w_1, w_2, .., w_m, b_1, b_2, b_3, .., b_k\} \qquad (23)$$

Eq. (23) shows the set of weights $\{w_1, w_2, w_3, \ldots, we_m\}$ and the set of biases $\{b_1, b_2, b_3, \ldots, b_k\}$ of the MLP coded as Harris Hawk or a solution in the swarm of HHO. In the HHO-MLP, the set of solutions that consist of weights and biases composes the population or the swarm of the HHO algorithm. These candidate solutions are initiated randomly in the search space. Eq. (24) shows the HHO swarm.

$$S = \{S(1), S(2), S(3), \ldots, S(n)\} \qquad (24)$$

S is the population or swarm of Harris hawks and n is the population size. Furthermore, S represents the possible MLPs. Each solution or MLP inside the population needs to be evaluated to determine its goodness. Eq. (25) shows the fitness function which represents the intrusion detection error rate.

$$\text{Err} = \frac{1}{n}\left(\sum_{i=1} (\hat{E}_i - E_i)^2\right) \qquad (25)$$

where $E_i$ and $\hat{E}_i$ are the actual and the detected classes of the $i_{th}$ network traffic respectively. The next step is to update the weights and biases in each vector of the HHO population. The update mechanism is applied in each iteration to select the optimal weights and biases and minimize the intrusion detection error by MLP. In algorithm 2, The algorithm's pseudo-code for the HHO-MLP is displayed.

## 5. The experiment and results

### 5.1. Description of the Network Dataset

The HHO-MLP method is evaluated using the KDD dataset. It is a collection of network traffic from a single host and every other node in the network. The KDD dataset is comprised of 42 features (34 numerical and 8 non-numerical), and the IDS is evaluated using a specific portion of its occurrences. [35]. This is because it compromises normal and suspicious traffic which makes it promising for evaluating intrusion detection. Furthermore, it doesn't limit network traffic in real-time and it compromises four types of intrusions: U2R, R2L, DOS, and Prob. The KDDCUP99 training dataset distribution is done as in the following (#instances, ratio): Normal (97277, 19.69%), DOS (391458, 79.24%), Probe (4107,0.83%), R2L (1126, 0.23%), U2L(52, 0.01%), so that the total (494,019, 100%). Regarding the UNSW-NB15 dataset, it simulates nine different types of attacks. The attacks include DOS, ShellCode, Worms, Fuzzers, Backdoors, Exploits, Analysis, Generic, and Reconnaissance.

Implementation of the proposed HHO-MLP is using the EvoloPy-NN open-source python framework [1]. It consists of a set of hybrid evolutionary algorithms integrated with ANN. The HHO has a population of 10 and has gone through 30 iterations. The neural network has 2 and 5 hidden layers and neurons, respectively. respectively.

### 5.2. IDS Security Evaluation Measures

The accuracy, sensitivity, and specificity [9] measures are used to evaluate the proposed HHO-MLP approach to detect network intrusions. These are shown in Eq. (26) to Eq. (27). True Positive intrusion detection (TP-ID), False Positive intrusion detection (FP-ID), True Negative intrusion detection

---
[1]https://github.com/7ossam81/EvoloPy-NN

---
**Algorithm 2** HHO-MLP pseudo-code
---
**Inputs**: The swarm size N and the number of all cycles T
**Coding**: Each harris hawk population: $S_1(i) = \{w_1, w_2, .., w_m, b_1, b_2, b_3, .., b_k\}$
**Outputs**: The position of prey and its fitness value
Initialize the random swarm Solution$_i(i = 1, 2, \ldots, N)$
**while** (the number of cycles is less than T) **do**
    MLP training with any weight and bias of solutions
    Compute the fitness values of solutions using the following equation
    $error = \frac{1}{n}(\sum_{i=1}^{n}(\hat{E}_i - E_i)^2)$
    Set $S_{prey}$ as the position of prey. (The best weights and biases vector)
    **for** (each ($S_i$)) **do**
        Update $E_0$ and J         ▷ $E_0$=2random()-1, J=2(1-random())
        Update the Energy using:
    Energy = $2Energy_0(1 - \frac{iter}{T})$
    **if** (Energy ≥ 1) **then**
        Update the solution position according to the following equation:

$$S(iter+1) = \begin{cases} S_{random}(iter) - r_1 S_{random}(iter) - 2r_2 S(iter) &, q \geq 0.5 \\ S_{prey}(iter) - S_m(iter) - r_3(LoBo + r_4(UpBo - LoBo)) &, q < 0.5 \end{cases}.$$

    **if** (E < 1) **then**
        **if** (r ≥0.5 and E ≥ 0.5 ) **then**
            Update the hawk position using
            $S(iter + 1) = \Delta S(iter) - EJS\ prey(iter) - S(iter)$
        **else if** (r≥ 0.5 and E < 0.5 ) **then**
            Update the hawk position using
            $S(iter + 1) = S_{prey}(iter) - E\Delta S(iter)$
        **else if** (r <0.5 and E ≥ 0.5 ) **then**
            Update the hawk position using:
$$S(iter+1) = \begin{cases} Y & \text{if } F(Y) < F(S(iter)) \\ Z & \text{if } F(Z) < F(S(iter)) \end{cases}$$
            where Y and Z are obtained using:
            $Y = S_{prey}(iter) - EJS_{(prey}(iter) - S(iter)$
The LeFl is applied as in Eq. (8) $Z = Y + Size \times LeFl(Dim)$
        **else if** (r <0.5 and E < 0.5 ) **then**
            Update the hawk position using:
$$S(iter+1) = \begin{cases} Y & \text{if } F(Y) < F(S(iter)) \\ Z & \text{if } F(Z) < F(S(iter)) \end{cases}.$$
$Y = S_{prey}(iter) - EJS_{prey}(iter) - S_m(iter)$
$Z = Y + Size \times LeFl(Dim)$
**Return** $S_{prey}$ with best weights and biases.
MLP training using the best weights and biases vector ($S_{prey}$)
---

(TN-ID), and False Negative intrusion detection (FN-ID) are needed for these calculations. As shown by the equations Eq. (29) and Eq. (30), other metrics used include mean average-error-index and squared-error when distinguishing legitimate from anomalous traffic detection errors are employed. Eq. (26) shows the accuracy of the intrusion detection system.

$$\text{ID-acc} = \frac{TP\text{-}ID + TN\text{-}ID}{TP\text{-}ID + TN\text{-}ID + TP\text{-}ID + FN\text{-}ID} \qquad (26)$$

Eq. (27) shows the specificity of the intrusion detection system.

$$\text{ID-spec} = \frac{TN\text{-}ID}{TN\text{-}ID + FP\text{-}ID} \qquad (27)$$

Eq. (28) shows the sensitivity of the intrusion detection system.

$$\text{ID-sens} = \frac{TP\text{-}ID}{TP\text{-}ID + FN\text{-}ID} \qquad (28)$$

Eq. (29) shows the average-error-index of the intrusion detection system.

$$\text{ID-mse} = \frac{1}{n} \left( \sum_{i=1}^{n} (\hat{E}_i - E_i)^2 \right) \qquad (29)$$

Eq. (30) shows the squared-error of the intrusion detection system.

$$\text{ID-rmse} = \sqrt{\frac{1}{n} \left( \sum_{i=1}^{n} (\hat{E}_i - E_i)^2 \right)} \qquad (30)$$

### 5.3. Discussion and limitations

Fig. 3 shows the results of different experiments when 30 iterations and initial swarm sizes of 5, 10, 15, 20, and 30 are considered. The results prove the effect of swarm size on reducing the network IDS error rate. As it is clearly shown that the MSE decreases dramatically from 0.205 to 0.086. The error rate decreases in ascending order relative to the swarm size. This means that the minimum error rate is achieved for the largest swarm size. In addition, regardless of the starting swarm size, the increased iteration count helps the MLP converge to the best weights and biases, which in turn lowers the error rate.

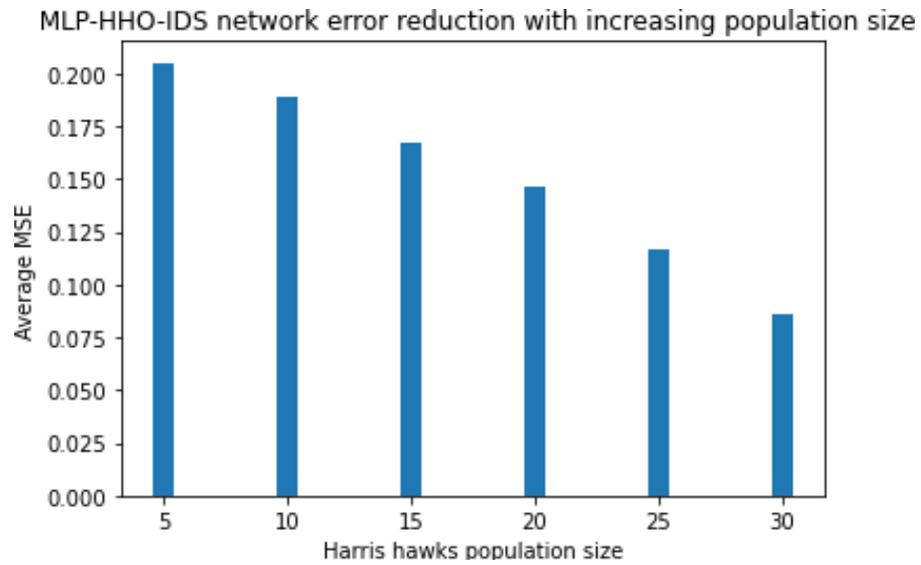

Figure 3: Inverse relationship between the Harris Hawks swarm size and the intrusion detection error rate in a network.

Fig 4 shows the values of Intro-detect-accuracy, Intro-detect-sensitivity, and Intro-detect-specificity of SVM, PSO-C4.5-IDS, PSO-SVM, HHO-MLP, MLP [36].

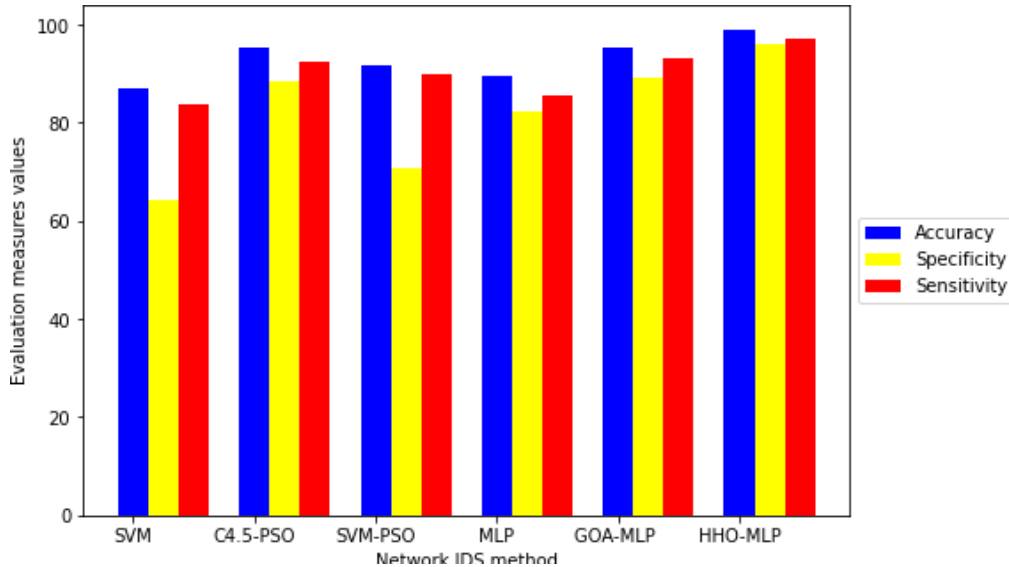

Figure 4: Comparison of HHO-MLP against other methods in the literature on the KDD dataset in terms of sensitivity and accuracy.

Notice that in this comparison we try to study the efficiency of the new model against other hybrid methods in which other evolutionary algorithms are integrated with the MLP algorithm. Furthermore, it is compared with other learning algorithms implemented without hybridization with evolutionary algorithms. Comparative analysis indicates that the Hybrid HHO-MLP outshone other methods when assessed using selected evaluation metrics. HHO-MLP showed superior performance by attaining top scores in accuracy, sensitivity, and specificity measurements; specifically, an accuracy rate of 93.17%, sensitivity level of 89.25%, and specificity percentage of 95.41% were recorded by this approach. It appears that the hybridization of the HHO algorithm with MLP enhances the intrusion detection rate. The HHO-MLP achieved higher accuracy, sensitivity, and specificity compared with the standard MLP by 4.77%, 6.89%, and 7.49%, respectively. In Fig 5, we observe a comparative analysis between the suggested Hybrid HHO-MLP technique and other well-established algorithms, such as Binary Particle Swarm Optimization (Bi-PSO), Binary Bat Algorithm incorporating Levy Flights (Bi-BA-LF), Binary Firefly Algorithm (Bi-FA), and the Naive Bayes (NB) classifier [37], [38].

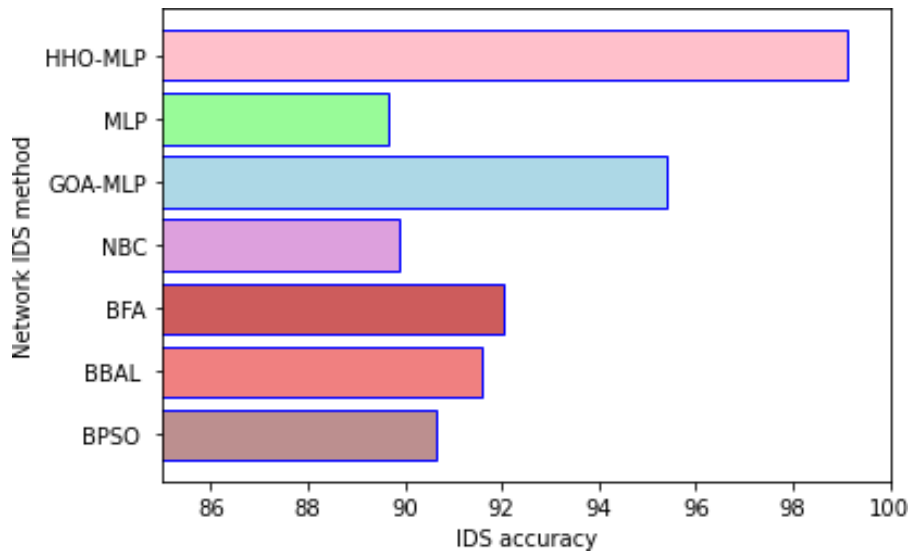

Figure 5: Comparison of HHO-MLP against other methods in the literature based on accuracy.

The next experiments conduct comparisons between the proposed HHO-MLP and other hybrid evolutionary algorithms that are integrated with the MLP learning algorithm. They used evolutionary algorithms including BOA, GOA, and BOW. Furthermore, RF and XGBoost are considered for the validation of the proposed approach. Evaluations of these algorithms were carried out utilizing two widely recognized datasets, KDD Cup 1999 and UNSW-NB15, with the primary goal of gauging their ability to detect cyber intrusions. This approach provides a comprehensive comparison between techniques in terms of their detection performance. Tables 1 and 2, show these algorithm. The results show that HHO-MLP has achieved the highest accuracy in detecting intrusions in networks. Then, the GOA-MLP comes in the next place. XGBoost has the third order in terms of accuracy. Furthermore, the proposed HHO-MLP has the highest specificity, while GOA-MLP and BWO-MLP are in third place.

Table 1: Comparison between HHO-MLP and other hybrid evolutionary algorithms with MLP over the KDD dataset.

| IDS name | ID-acc | ID-sens | ID-spec | ID-time (second) |
|---|---|---|---|---|
| HHO-MLP | 99.13 | 96.10 | 97.29 | 0.2 |
| BOA-MLP | 93.82 | 93.29 | 93.54 | 0.4 |
| GOA-MLP | 95.41 | 93.17 | 89.25 | 0.3 |
| BWO-MLP | 94.66 | 94.26 | 94.33 | 0.8 |
| RF | 94.23 | 93.92 | 94.19 | 1.2 |
| XGBoost | 95.15 | 94.82 | 94.88 | 1.5 |

Table 2: Comparison between HHO-MLP and other hybrid evolutionary algorithms with MLP over the UNSW dataset.

| IDS name | ID-acc | ID-sens | ID-spec | ID-time (second) |
|---|---|---|---|---|
| HHO-MLP | 99.23 | 98.34 | 98.45 | 0.1 |
| BOA-MLP | 97.83 | 97.44 | 97.58 | 0.3 |
| GOA-MLP | 98.88 | 98.09 | 98.14 | 0.2 |
| BWO-MLP | 98.19 | 98.07 | 98.12 | 0.7 |
| RF | 97.82 | 97.59 | 97.63 | 0.9 |
| XGBoost | 98.33 | 97.66 | 98.08 | 1.2 |

It is worth this regard to point to the essential criteria of the network IDS which is the intrusion detection speed. An additional benefit of the proposed HHO-MLP is that it performs the training process on a small part of the network traffic. Then, a feature selection is applied by filtering some significant features of the network traffic. This reduced the size of the dataset from 42 to 15 feature subsets. Conducting a training process on a smaller size of features helped to speed up the process so implementing the HHO-MLP takes less time than the standard MLP. The main limitation of the proposed HHO-MLP is that it was not validated under real network environment to check the robustness of the approach and to check its capability to detect unkown types of threats

## 6. Conclusion

This study presents a novel network-intrusion detector system. The Multilayer Perceptron learning algorithm (MLP), combined with the Harris Hawks optimization technique (HHO), reduces the network intrusion error. This hybrid HHO-MLP system aims to optimize MLP parameters during the learning process to reduce network detection errors.

This methodology has been implemented within the EvoloPy-NN Python framework and evaluated using KDD and UNSW datasets, with its performance measured against several well-established algorithms including RF, XGBoost, ANN), BOA, GOA, and BWO. The HHO-MLP model showed its superiority across three evaluation metrics by reaching 93.17% accuracy, 95.25% sensitivity, and 95.41% specificity respectively.

Future research endeavors aim to explore multi-objective swarm intelligence algorithms combined with FS techniques to optimize classification methods across various network infrastructures. Furthermore, we intend to implement other evolutionary algorithms with deep learning methodologies to enhance intrusion detection capabilities across varying networking systems such as internet of things (IoT).

**Acknowledgements**

This work was supported by the Ministerio Español de Ciencia e Innovación under project number PID2020-115570GB-C22 MCIN/AEI/10.13039/501100011033 and by the Cátedra de Empresa Tecnología para las Personas (UGR-Fujitsu).